\documentclass{article}
\usepackage{spconf,amsmath,graphicx}
\usepackage{booktabs}
\usepackage{xcolor}
\usepackage{url}

\newcommand{\tx}{t_x}
\newcommand{\ty}{t_y}

\title{Fast Slice-Wise Deep Learning Motion Correction for Spinal Cord Functional MRI}

\makeatletter
\gdef\@name{{\em Thomas Dagonneau$^{1,2}$, Sandrine B\'edard$^{1}$, Nontharat Tucksinapinunchai$^{3}$, Uten Yarach$^{3}$,}\\
{\em Lo\"ic Le Folgoc$^{2}$, Pietro Gori$^{2}$, Julien Cohen-Adad$^{1}$}\\}
\makeatother
\address{$^{1}$Polytechnique Montr\'eal, Montr\'eal, QC, Canada \\
$^{2}$T\'el\'ecom Paris, Institut Polytechnique de Paris, Palaiseau, France \\
$^{3}$Faculty of Associated Medical Sciences, Chiang Mai University, Chiang Mai, Thailand}

\begin{document}
\maketitle

\begin{abstract}
Respiratory and bulk motion displace the spinal cord between the volumes of a
functional MRI timeseries. Current methods take several minutes per time series, and general-purpose brain tools register
whole volumes, whereas the cord moves mainly slice by slice. We propose a lightweight
convolutional neural network that regresses the in-plane translation aligning
each axial slice to a reference. Two models, trained without motion labels,
correct large and then residual motion. On a 3T (120 scans) and a 7T (59 scans) dataset, the proposed method improves on SCT on every metric
(tSNR $+10$ to $+25\%$, DVARS $-20$ to $-38\%$), leaves almost no slice
displaced by more than one voxel ($<0.1\%$), performs on par with a slice-wise AFNI
variant and better than slice-wise FSL FLIRT, and
runs in about 10\,s per scan, about 30 times faster than Spinal Cord Toolbox (SCT). The code is available at
\url{https://github.com/ivadomed/moco-dl} and the method is available in SCT v7.4.
\end{abstract}


\section{Introduction}
\label{sec:intro}

In spinal cord functional MRI, respiration and voluntary movements displace
the cord between volumes~\cite{fratini2014}, which lowers the temporal
signal-to-noise ratio (tSNR) and adds fluctuations that can hamper the
detection of functional activity. Unlike the brain, the cord does not move as a rigid block: its
displacement varies along its length, mostly within the axial plane, and the
thick axial slices (3--5\,mm) are acquired at different times, so motion
is usually corrected slice by slice with in-plane
transforms~\cite{cohenadad2009slicecorr,deleener2017sct}, as we do here.

The Spinal Cord Toolbox (SCT, v7.3)~\cite{deleener2017sct} provides a slice-wise
correction, \texttt{sct\_fmri\_moco}, built on ANTs~\cite{avants2011ants}, which
takes 4--7\,min per time series. Brain tools such as AFNI
\texttt{3dvolreg}~\cite{cox1996afni} and FSL MCFLIRT~\cite{jenkinson2002flirt}
register whole volumes and must be adapted to slice-wise
registration around the cord, as in the Neptune toolbox~\cite{neptune2022}.
DeepRetroMoCo~\cite{mobarak2024deepretromoco}, an unsupervised network
predicting dense deformation fields, was trained and tested on a single 3T
dataset and is released as research code rather than a ready-to-use tool.

Our main contribution is a fast, ready-to-use tool, available in SCT v7.4: a
lightweight CNN regressing slice-wise in-plane translations, trained without
labels on multi-site 3T and 7T data with random-walk motion augmentation, and
applied in two stages (large, then residual motion) with an NCC check that
rejects harmful predictions. We compare it with
\texttt{sct\_fmri\_moco} and slice-wise AFNI and FSL FLIRT on two cord fMRI
datasets, at 3T and 7T.

\begin{figure*}[t]
  \centering
  \includegraphics[width=\textwidth]{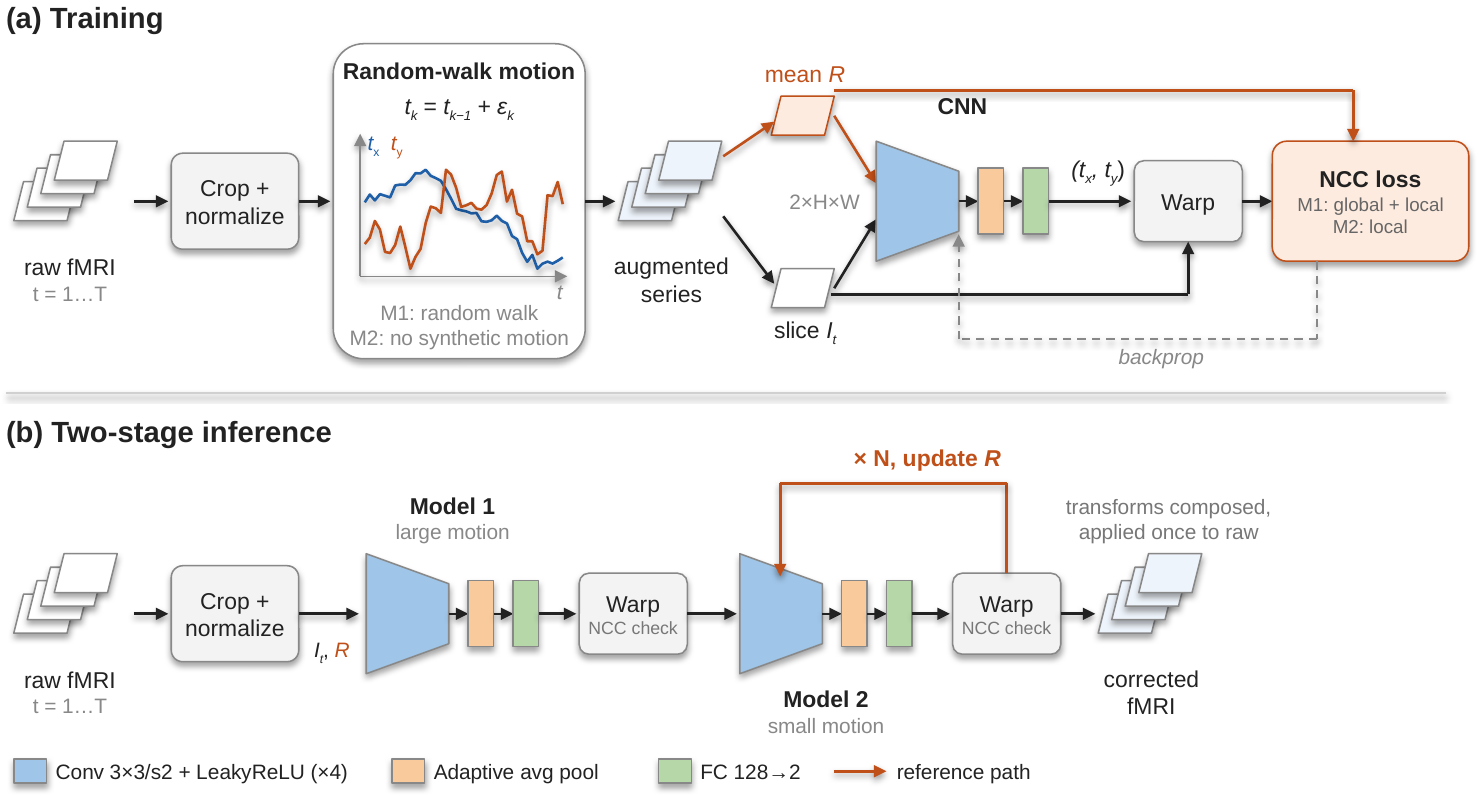}
  \caption{Overview. (a)~Training: the CNN predicts $(\tx,\ty)$ from a slice
  $I_t$ and the temporal mean $R$; an NCC loss compares the warped slice with
  $R$. M1 sees random-walk displacements of up to 3 voxels. (b)~Inference: M1,
  then M2 up to $N=6$ times, updating $R$; translations that lower the NCC are
  discarded and the composed transform is applied once to the raw data.}
  \label{fig:pipeline}
\end{figure*}

\section{Methods}
\label{sec:methods}

\subsection{Network and preprocessing}
For each axial slice of each volume, the network takes the slice to correct,
$I_t$, and a reference slice, $R$ (by default the temporal mean), stacked as
channels, and outputs the translation $(\tx,\ty)$ aligning $I_t$ to $R$
(Fig.~\ref{fig:pipeline}). The architecture consists of four $3{\times}3$ stride-2 convolutions with
LeakyReLU (16 to 128 channels), adaptive average pooling and a linear layer
for about 98k parameters in total. Images are first cropped around the cord with
\texttt{sc-crop} (detection on the first volume, same box for all volumes), which removes
background structures that move differently from the cord and speeds up
inference.
Each slice is padded or cropped to $64{\times}64$ and its intensities are
clipped to the 1\textsuperscript{st}--99\textsuperscript{th} percentiles and rescaled to $[0,1]$.

\subsection{Training}
Two models share this architecture: M1 for large, persistent movements and M2
for small residual ones. They are trained on five public OpenNeuro datasets:
ds004386, ds004616, ds005075, ds006729, ds007932, comprising 141 subjects total acquired at 3T (n=133) and 7T (n=8), in-plane resolution 0.75--3.4\,mm, slice thickness 1--5\,mm, TR = 1.55--3.2\,s. The method is tested on two other public datasets, unseen
during training (ds008784 and ds007067, Section~\ref{sec:data}). Some of the datasets included multiple runs per subject: the total number of runs was 420, which were split by
subject into 337 for training and 83 for validation, giving 2{,}603{,}976 and
683{,}041 slice pairs.

For M1, a synthetic random-walk translation is added to each series:
$t_k = t_{k-1} + \varepsilon_k$, $\varepsilon_k \sim \mathcal{N}(0,\sigma^2)$,
independently along $x$ and $y$, with $\sigma=3$ voxels and each component
clipped to $[-3,3]$ voxels, so that displacements drift and persist as real
motion does.  M2 is trained on
the real series without synthetic motion, to focus on the small displacements
present in the data. The loss compares the warped slice with the mean $R$ of the series after data augmentation. For M1, the loss
$\mathcal{L}_{1} = \mathcal{L}_{g} + \mathcal{L}_{l}$ combines a global NCC over
the non-zero pixels, $\mathcal{L}_{g}$, which keeps a useful gradient for large
displacements, with a local NCC ($5{\times}5$ window~\cite{monai}),
$\mathcal{L}_{l}$, sensitive to sub-voxel misalignment. For M2,
$\mathcal{L}_{2} = \mathcal{L}_{l}$ uses the local NCC only. Both models are trained with Adam (learning rate $10^{-4}$, 100
epochs, batch size 128), keeping the checkpoint with the
lowest validation loss on real pairs; each takes about 13\,h on an NVIDIA RTX
PRO 6000 GPU.

\subsection{Two-stage iterative inference}
M1 predicts a translation for every slice, with $R$ the temporal mean; any
translation that lowers the NCC between the slice and $R$ is replaced by the
identity. M2 then refines every slice iteratively: $R$ is recomputed from the
corrected series, and M2 predicts a residual translation, filtered by the same
check and composed with the current transform. Refinement stops after $N=6$
iterations, when the mean parameter change falls below $10^{-3}$, or when it
increases (the previous transform is then kept). The composed transform is
applied once to the raw, uncropped slices.

\begin{figure}[t]
  \centering
  \includegraphics[width=0.92\columnwidth]{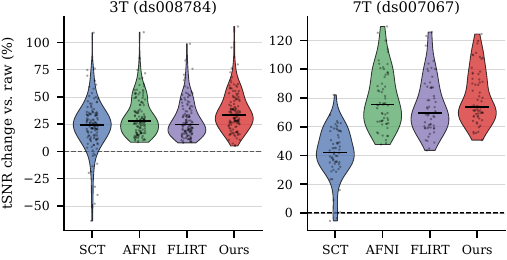}
  \caption{Per-scan tSNR change relative to the raw data, bilinear
  interpolation; bars: medians. Ours improves tSNR in every scan, whereas
  \texttt{sct\_fmri\_moco} lowers it in some.}
  \label{fig:tsnr}
\end{figure}

\begin{figure*}[t]
  \centering
  \includegraphics[width=0.86\textwidth]{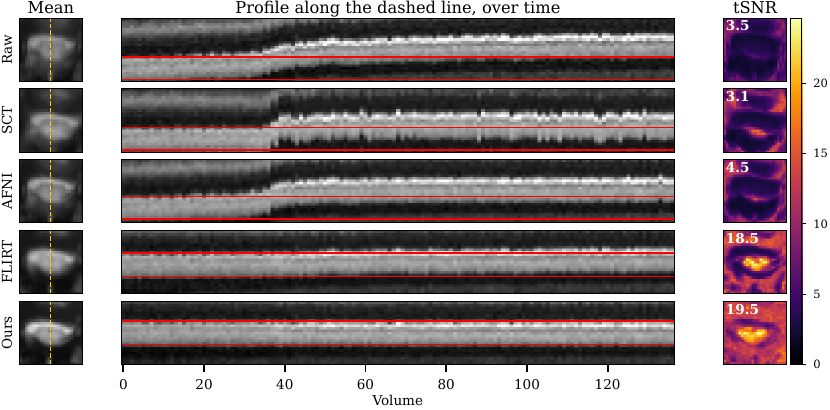}
  \caption{Example scan with severe motion, one slice: temporal mean, intensity along
  the dashed line over time, and tSNR map (mean tSNR in the cord of this
  slice). Red lines: spinal cord edges in the first volumes.}
  \label{fig:kymo}
\end{figure*}
\section{Experiments and Results}
\label{sec:results}

\subsection{Datasets, baselines and metrics}
\label{sec:data}
Both test datasets are public on OpenNeuro. {ds008784}~\cite{bedard2026}
contains 120 runs (40 healthy subjects, 3 runs each) acquired at 3T
($1.25{\times}1.25$\,mm in-plane, 4--5\,mm slices, TR = 1.7\,s).
{ds007067}~\cite{ds007067} contains 59 runs from 14 subjects at 7T
($0.75{\times}0.75$\,mm in-plane, 3\,mm slices, TR = 2.5\,s).

We compare against \texttt{sct\_fmri\_moco} (SCT v7.3) and a slice-wise AFNI
variant following Neptune~\cite{neptune2022} (hereafter AFNI), in which
\texttt{3dWarpDrive} estimates the in-plane translation of each axial slice. We also apply FSL
FLIRT~\cite{jenkinson2002flirt} slice by slice (2D rigid, weighted by a cord
mask), the approach of expert spinal cord pipelines: it estimates in-plane
rotation in addition to translation, but optimizes each slice separately. All
methods align to the temporal mean and, as interpolation reduces noise and changes the metrics by
itself, use bilinear interpolation. 
Whole-volume tools such as
\texttt{3dvolreg} are not included: they apply one rigid transform per volume,
whereas the cord moves differently at each level and each slice is acquired
at a different time.

With no ground truth available, we use four metrics within a cord mask (the
ds008784 segmentation, otherwise \texttt{sct\_deepseg} \texttt{sc\_epi} on the raw
temporal mean), one value per
scan: tSNR; DVARS (RMS intensity change between consecutive volumes, in \% of
the mean signal); sharpness (mean gradient magnitude of the temporal mean near
the cord over its mean intensity), which drops with residual motion and blur;
and outlier slices, the percentage of slices whose cord is still displaced by
more than one voxel from the temporal mean (cross-correlation around the cord).
Times exclude file input/output, on the CPU of one workstation (Intel Xeon
w7-2495X). Methods are compared with ours by paired Wilcoxon tests,
Holm-corrected within each metric.

\subsection{Results}

\begin{table}[t]
  \centering
  \caption{Results on ds008784 (3T, 120 scans, mean over scans, bilinear
  interpolation). Outl.: slices displaced by $>1$ voxel (\%); time per scan.
  Bold: best. $^{*}$: different from ours (paired Wilcoxon tests, Holm-corrected $p<0.05$).}
  \label{tab:gt}
  \footnotesize\setlength{\tabcolsep}{2pt}
  \resizebox{\columnwidth}{!}{%
  \begin{tabular}{@{}lccccc@{}}
    \toprule
    Method & tSNR$\uparrow$ & DVARS$\downarrow$ & Sharp.$\uparrow$ & Outl.$\downarrow$ & Time (s) \\
    \midrule
    Raw                      & 14.39$^{*}$ & 9.67$^{*}$ & 0.177$^{*}$ & 0.788$^{*}$ & -- \\
    \midrule
    \texttt{sct\_fmri\_moco} & 17.65$^{*}$ & 9.98$^{*}$ & 0.164$^{*}$ & 2.347$^{*}$ & 251.8$^{*}$ \\
    AFNI                   & 18.74$^{*}$ & 8.18$^{*}$ & 0.171$^{*}$ & 0.041$^{*}$ & 18.4$^{*}$ \\
    FSL FLIRT              & 18.40$^{*}$ & 8.23$^{*}$ & \textbf{0.173}$^{*}$ & 0.010$^{*}$ & 256.5$^{*}$ \\
    Ours                     & \textbf{19.34} & \textbf{7.96} & 0.170 & \textbf{0.003} & \textbf{10.4} \\
    \bottomrule
  \end{tabular}}
\end{table}

\begin{table}[t]
  \centering
  \caption{Results on ds007067 (7T, 59 scans), as in Table~\ref{tab:gt}.}
  \label{tab:7t}
  \footnotesize\setlength{\tabcolsep}{2pt}
  \resizebox{\columnwidth}{!}{%
  \begin{tabular}{@{}lccccc@{}}
    \toprule
    Method & tSNR$\uparrow$ & DVARS$\downarrow$ & Sharp.$\uparrow$ & Outl.$\downarrow$ & Time (s) \\
    \midrule
    Raw                      & 6.87$^{*}$ & 26.31$^{*}$ & 0.170$^{*}$ & 15.62$^{*}$ & -- \\
    \midrule
    \texttt{sct\_fmri\_moco} & 9.76$^{*}$ & 20.17$^{*}$ & 0.170$^{*}$ & 4.73$^{*}$ & 405.0$^{*}$ \\
    AFNI                   & 12.09 & \textbf{12.23} & \textbf{0.196}$^{*}$ & 0.12 & 32.4$^{*}$ \\
    FSL FLIRT              & 11.78$^{*}$ & 12.63 & 0.193 & 0.21$^{*}$ & 351.1$^{*}$ \\
    Ours                     & \textbf{12.21} & 12.40 & 0.193 & \textbf{0.06} & \textbf{10.7} \\
    \bottomrule
  \end{tabular}}
\end{table}

\textit{3T data.} With bilinear interpolation
(Table~\ref{tab:gt}), our method improves on \texttt{sct\_fmri\_moco} on every
metric (all $p<10^{-12}$) and is 24 times faster. It
improves tSNR in every scan, whereas \texttt{sct\_fmri\_moco} lowers it in
some (Fig.~\ref{fig:tsnr}). It improves on
AFNI on every metric but sharpness ($p<10^{-7}$). It also improves on FLIRT in tSNR, DVARS and outlier
slices ($p<0.02$) and is 25 times faster; FLIRT, which also estimates
rotation, keeps slightly sharper images (0.173 vs.\ 0.170).

\textit{7T data.} With bilinear interpolation (Table~\ref{tab:7t}), our method
improves on \texttt{sct\_fmri\_moco} on every metric (tSNR 12.21 vs.\ 9.76,
DVARS 12.40 vs.\ 20.17, all $p<10^{-9}$) and is 38 times faster. Outlier
slices drop from 15.6\% before correction to 0.06\%, against 4.7\%.
AFNI matches our tSNR and outlier slices, with slightly better DVARS and
sharpness, but is three times slower. FLIRT has a lower tSNR and more outlier
slices (0.21\%, $p=0.001$), with similar DVARS and sharpness, and is 33 times
slower (351\,s).

\textit{Severe motion.} Fig.~\ref{fig:kymo} shows an example scan (40\% of slices
displaced) in which the cord jumps by several voxels at volume 37. Only our method keeps the spinal cord
within its initial edges for the whole scan: FLIRT reduces the
jump but leaves the cord shifted by about one voxel afterwards, and
\texttt{sct\_fmri\_moco} and AFNI leave it displaced. Run-level metrics
barely reflect this: AFNI still raises the tSNR from 4.4 to 7.7 despite
27.9\% outlier slices (0\% for ours).


\section{Discussion and Conclusion}
\label{sec:conclusion}

Compared with \texttt{sct\_fmri\_moco}, a small unsupervised CNN with 98k
parameters corrects spinal cord motion in fMRI timeseries better and runs 24
to 38 times faster. It performs on par with AFNI and better than slice-wise FLIRT, although more extensive tests are warranted.

The main limitation is the evaluation: with no ground truth, and all methods
aligning to the temporal mean, the metrics measure temporal consistency rather
than anatomical accuracy and depend on the interpolation. Simulated motion of
known amplitude would quantify accuracy directly. Also, the network estimates
translations only, whereas FLIRT also corrects in-plane rotation, which may explain its slightly sharper images at 3T. Future work will further validate in other databases, and extend the
method to spinal cord diffusion MRI.

\section{Compliance with Ethical Standards}
\label{sec:ethics}
This retrospective study uses open-access human data from OpenNeuro;
ethical approval was not required, as confirmed by the data license.

\section{Acknowledgments}
\label{sec:acknowledgments}
This work was funded by AMX (France), CIHR, NSERC, Praxis and NVIDIA. The authors have no relevant financial or
non-financial interests to disclose. Claude (Anthropic) assisted with
writing and code; the authors reviewed all content.

\bibliographystyle{IEEEbib}
\bibliography{refs}

@article{fratini2014,
  author  = {Fratini, M. and Moraschi, M. and Maraviglia, B. and Giove, F.},
  title   = {On the impact of physiological noise in spinal cord functional {MRI}},
  journal = {J. Magn. Reson. Imaging},
  volume  = {40},
  number  = {4},
  pages   = {770--777},
  month   = oct,
  year    = {2014}
}

@article{deleener2017sct,
  author  = {De Leener, B. and others},
  title   = {{SCT}: {Spinal Cord Toolbox}, an open-source software for processing spinal cord {MRI} data},
  journal = {NeuroImage},
  volume  = {145, Part A},
  pages   = {24--43},
  year    = {2017}
}

@article{avants2011ants,
  author  = {Avants, B. B. and Tustison, N. J. and Song, G. and Cook, P. A. and Klein, A. and Gee, J. C.},
  title   = {A reproducible evaluation of {ANTs} similarity metric performance in brain image registration},
  journal = {NeuroImage},
  volume  = {54},
  number  = {3},
  pages   = {2033--2044},
  year    = {2011}
}

@article{cox1996afni,
  author  = {Cox, R. W.},
  title   = {{AFNI}: software for analysis and visualization of functional magnetic resonance neuroimages},
  journal = {Comput. Biomed. Res.},
  volume  = {29},
  number  = {3},
  pages   = {162--173},
  year    = {1996}
}

@article{jenkinson2002flirt,
  author  = {Jenkinson, M. and Bannister, P. and Brady, M. and Smith, S.},
  title   = {Improved optimization for the robust and accurate linear registration and motion correction of brain images},
  journal = {NeuroImage},
  volume  = {17},
  number  = {2},
  pages   = {825--841},
  year    = {2002}
}

@article{mobarak2024deepretromoco,
  author  = {Mobarak-Abadi, M. and others},
  title   = {{DeepRetroMoCo}: deep neural network-based retrospective motion correction algorithm for spinal cord functional {MRI}},
  journal = {Front. Psychiatry},
  volume  = {15},
  pages   = {1323109},
  month   = jun,
  year    = {2024}
}

@article{monai,
  author  = {Cardoso, M. J. and others},
  title   = {{MONAI}: An open-source framework for deep learning in healthcare},
  journal = {arXiv preprint arXiv:2211.02701},
  year    = {2022}
}

@article{bedard2026,
  author  = {B{\'e}dard, S. and others},
  title   = {Spatial distribution of spinal cord {fMRI} activity with electrocutaneous stimulation},
  journal = {Imaging Neuroscience},
  volume  = {4},
  pages   = {IMAG.a.1383},
  doi     = {10.1162/IMAG.a.1383},
  year    = {2026}
}

@inproceedings{neptune2022,
  author    = {Rangaprakash, D. and Barry, R. L.},
  title     = {Neptune: a toolbox for spinal cord functional {MRI} data processing and quality assurance},
  booktitle = {Proc. Intl. Soc. Mag. Reson. Med. (ISMRM)},
  pages     = {0396},
  year      = {2022}
}

@misc{ds008784,
  author = {B{\'e}dard, S. and Kaptan, M. and Weber, K. A.},
  title  = {Dermatomal\_Mapping\_R01\_Aim1},
  howpublished = {OpenNeuro, doi:10.18112/openneuro.ds008784.v1.0.0},
  year   = {2026}
}

@misc{ds007067,
  author = {D'Astous, A. and others},
  title  =  {{Spine fMRI 7T}},
  howpublished = {OpenNeuro, doi:10.18112/openneuro.ds007067.v4.0.0},
  year   = {2026}
}

@inproceedings{cohenadad2009slicecorr,
  author    = {Cohen-Adad, J. and others},
  title     = {Slice-by-slice motion correction in spinal cord {fMRI}: {SliceCorr}},
  booktitle = {Proc. Intl. Soc. Mag. Reson. Med. (ISMRM)},
  volume    = {17},
  year      = {2009}
}

\end{document}